\pdfoutput=1
\documentclass[letterpaper]{article} 
\usepackage{aaai2027}  
\usepackage[hyphens]{url}  
\usepackage{graphicx} 
\usepackage{natbib}  
\usepackage{caption} 
\providecommand{\href}[2]{#2}
\usepackage[dvipsnames]{xcolor}

\usepackage{amsmath,amssymb}
\usepackage{tikz-cd}

\usepackage{algorithm}
\usepackage{algorithmic}

\usepackage{booktabs}
\usepackage{array}

\title{Do Activation Monitors Survive Model Updates? Benchmarking, Predicting, and Repairing Activation-Monitor Staleness}
\author{
    Evan Duan
}
\affiliations{
    University of Michigan\\
    evanduan@umich.edu
}

\begin{document}

\maketitle

\begin{abstract}

Activation monitors --- lightweight probes trained on a language model's internal representations --- are an increasingly common layer in deployment safety stacks. Deployed models however are rarely static: they are quantized, fine-tuned, adapted with LoRA, or served with merged adapters while the monitor remains frozen. We present the first systematic test of whether this implicit contract holds: whether activation monitors trained on a base model remain reliable after these routine model updates. Across multiple safety-relevant monitors, model depths, update families, and open-weight models, we find a sharp split: quantization-style updates largely preserve frozen probe performance, while fine-tuning-style updates frequently make probes stale. Fragility is highly monitor-dependent, with privacy/PII probes most affected and refusal-compliance probes comparatively stable, showing that retraining a behavior need not stale its corresponding monitor. QLoRA is especially damaging despite NF4 quantization alone being relatively benign, suggesting that quantization becomes riskier when combined with adaptation. We further show that degradation is predictable from pre-deployment features, enabling revalidation budgets to be triaged toward the monitors most likely to fail. Finally, we test repair strategies and find that cheap label-free activation realignment repairs every repair-relevant stale cell, with none requiring score calibration, few-label heads, or labeled retraining. These results suggest that fine-tuning should trigger activation-monitor revalidation by default, with prediction triaging which monitors to check first and label-free realignment as the default repair.

\end{abstract}

\begin{figure}[t]
\centering
\includegraphics[width=\columnwidth]{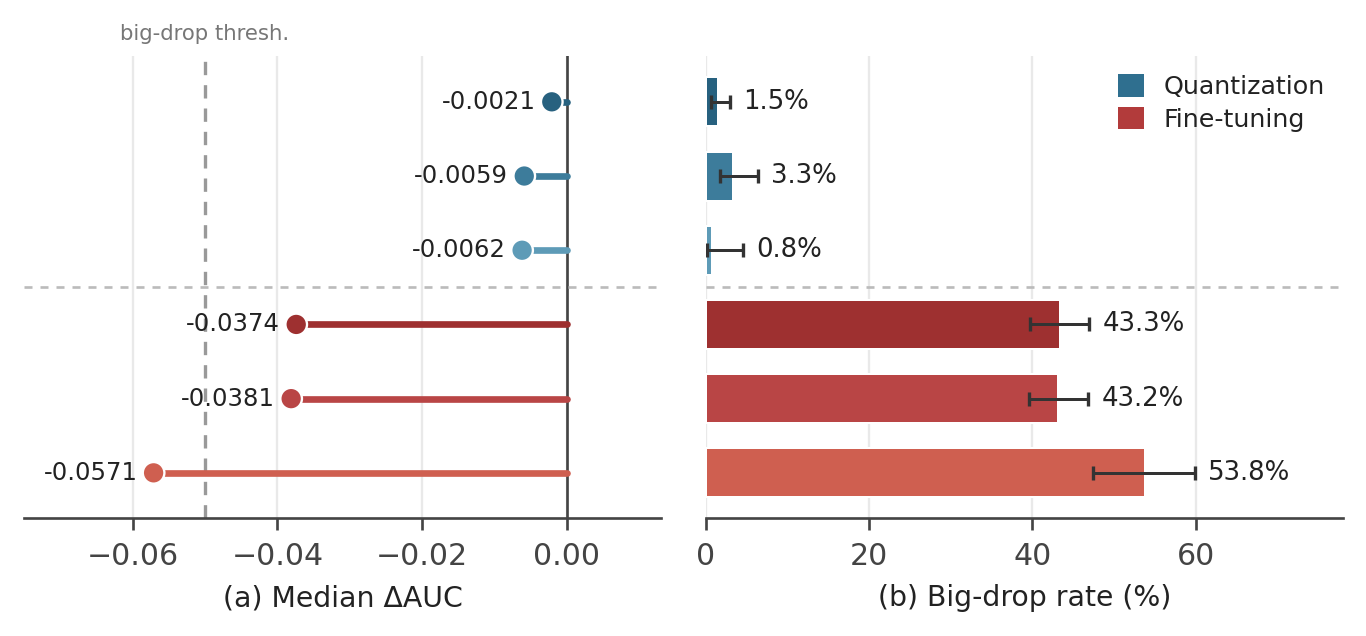}
\caption{\textbf{Update-family split in monitor staleness.} Median $\Delta$AUC and big-drop rates separate quantization-flat updates from fine-tune-hot updates. Dashed line: $\Delta$AUC $\leq -0.05$ big-drop threshold; error bars: Wilson 95\% CIs.}
\label{fig:update-family-split}
\end{figure}

\begin{figure*}[t]
\centering
\includegraphics[width=\textwidth,height=0.18\textheight]{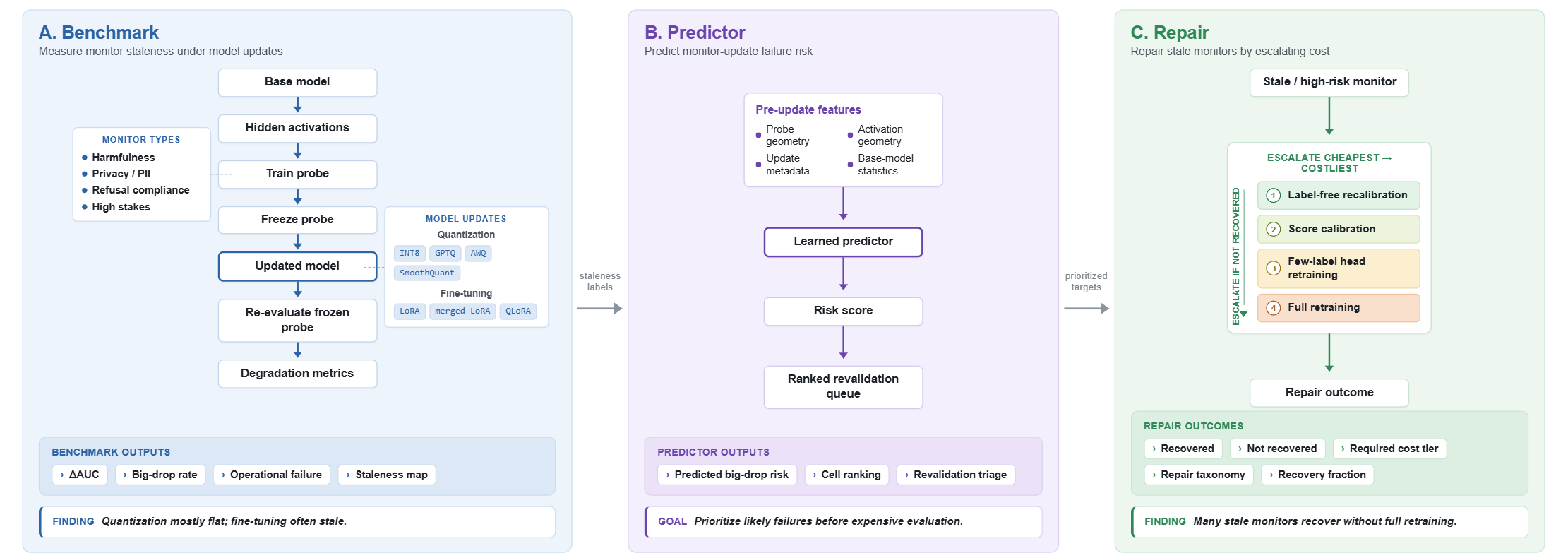}
\caption{\textbf{Overview of the activation-monitor maintenance pipeline.} We benchmark whether probes trained on base-model activations remain reliable after model updates by freezing the base probe and re-evaluating it on updated-model activations. The resulting staleness labels support both a predictor that prioritizes high-risk monitor--update pairs for revalidation and a cost-ordered repair ladder that escalates from label-free recalibration to full retraining only when needed.}
\label{fig:pipeline}
\end{figure*}

\section{Introduction}

Safety stacks for deployed language models increasingly include activation monitors: lightweight classifiers, often linear probes, trained on internal representations to flag safety-relevant conditions such as harmful requests, deceptive intent, or impending refusal \cite{zou2023representation,azaria2023internal,goldowskydill2025detecting}. Probes are cheaper than a second guard model, add little latency, and read internal model state rather than only sampled text, which output-level guardrails cannot do, thus its appeal \cite{inan2023llama,han2024wildguard}.

Deployed models, however, are moving targets. The same checkpoint may be quantized for cheaper serving, customized with parameter-efficient fine-tuning, served with runtime adapters, merged with those adapters, or adapted on an already-quantized base \cite{hu2022lora,dettmers2022llmint8,dettmers2023qlora,frantar2023gptqaccurateposttrainingquantization,lin2026awqactivationawareweightquantization,xiao2023smoothquant}. In each case, the deployed model changes while the activation monitor, trained once on base-model activations, remains frozen. The implicit contract is that the representation read by the probe remains stable under these updates. To our knowledge, this contract has not been systematically tested across routine post-deployment update families. 

We study: (i) which routine model-side updates degrade frozen activation monitors, and by how much; (ii) whether some monitoring targets are more fragile than others; (iii) whether degradation is predictable in advance from information available before post-update evaluation; and (iv) whether stale monitors can be cheaply repaired.

We answer these questions with a benchmark-and-maintenance pipeline. We train linear probes on base-model activations from Gemma-2-2B-it and Qwen2.5-7B-Instruct, freeze them, and re-evaluate them across twelve quantization, LoRA, merged-LoRA, and QLoRA update conditions, yielding 2,520 eligible update cells. We then test whether staleness can be predicted from pre-deployment metadata and base-side probe/activation features, and whether stale monitors can be repaired through a cost-ordered ladder from label-free activation realignment to labeled retraining.

Four findings emerge. First, fragility is strongly monitor-dependent: big-drop rates range from 57.1\% for privacy/PII to 7.9\% for refusal-compliance. Second, update family dominates staleness: quantization has low big-drop rates of 0.83–3.33\% and no operational failures, while LoRA-family updates produce 43.2–53.8\% big-drop rates and a 13.75\% operational-failure rate. Third, staleness is predictable before post-update evaluation: learned predictors beat the strongest lookup baseline in three of four leave-group-out settings, including leave-update-family-out by +0.136 Spearman. Fourth, stale monitors are repairable without new labels: in the targeted repair grid, label-free activation realignment repaired all 111 repair-relevant cells, with no cell requiring calibration, few-label heads, or retraining. Our contributions are: 

\begin{itemize}
    \item We systematically evaluate frozen safety probes under routine model-side updates across 2 models, 4 monitors, 12 update conditions, 6 layers, and 5 seeds, yielding 2,520 eligible update cells with fixed degradation criteria.
    \item We identify a structured staleness pattern: quantization is largely flat, fine-tuning is often hot, runtime and merged adapters degrade nearly identically, and monitor fragility varies by 7$\times$, with privacy/PII most fragile.
    \item We show that degradation is predictable before post-update evaluation: learned predictors using pre-deployment features outperform strong group-mean baselines, including under unseen update-family extrapolation.
    \item We show that label-free activation realignment repairs all repair-relevant stale cells in a complete 216-cell targeted grid, supporting recoverable representational drift rather than signal erasure.
\end{itemize}

\section{Related Work}

\paragraph{Activation probes and internal safety monitors.}
Probing classifiers are widely used to study neural representations \cite{alain2016understanding,belinkov2022probing}, and recent work uses them as safety-relevant readouts of LLM internals. Linear or low-complexity probes have been used to detect truthfulness, deception, sleeper-agent behavior, and other latent safety properties \cite{burns2023discovering,azaria2023internal,hubinger2024sleeper,goldowskydill2025detecting}. Representation-engineering work also shows that high-level behaviors can often be read or steered through activation-space directions \cite{zou2023representation,arditi2024refusal}. Recent safety work extends these ideas to safety-aware probing and activation-based intervention \cite{Wu2025Secure,Han2025SafeSwitch}. These studies show that activations contain useful safety signals, but they usually evaluate monitors on the same model or update regime for which they were developed. We instead ask whether a frozen activation monitor remains valid after the underlying model changes.

\paragraph{Text-level guardrails versus activation monitors.}
Text-level guardrails such as moderation systems, Llama Guard, WildGuard, and ShieldGemma
operate on prompts and outputs rather than hidden states
\cite{markov2023holistic,inan2023llama,han2024wildguard,zeng2024shieldgemma}.
This makes them easier to reuse across models, but prevents them from directly reading
pre-output internal states. Activation monitors provide that internal access, but are tied
to the representation geometry of a specific checkpoint. 

\paragraph{Model-side updates and safety degradation.}
Modern deployment pipelines often change models through quantization and
parameter-efficient fine-tuning. Quantization methods such as LLM.int8, NF4/QLoRA, GPTQ,
AWQ, and SmoothQuant reduce serving cost while preserving task performance
\cite{dettmers2022llmint8,dettmers2023qlora,frantar2023gptqaccurateposttrainingquantization,lin2026awqactivationawareweightquantization,xiao2023smoothquant},
while LoRA and QLoRA enable lightweight customization
\cite{hu2022lora,dettmers2023qlora}. These updates can also affect safety: fine-tuning can
compromise aligned models \cite{yang2023shadow,qi2024finetuning}, compression can change
trustworthiness and robustness \cite{hong2024compressed,egashira2024exploiting}, and
recent methods attempt to protect refusal directions, safety layers, or alignment-sensitive
subspaces \cite{Hsu2024Safe,Li2024Safety,Du2025Anchoring}.

\paragraph{Refusal representations, degradation prediction, and recent monitor work.}
Mechanistic studies suggest that refusal and safety behavior are organized in identifiable internal structures, including residual-stream directions, safety-sensitive layers, and sparse features \cite{arditi2024refusal,Li2024Safety,cunningham2024sparse,Yeo2025Understanding}. These structures can shift under tuning \cite{Du2025Anchoring}, and compression can damage feature fidelity even when perplexity remains stable \cite{duan2026perplexitymisssaefeature}. Related work predicts performance shifts or intervention side effects from pre-shift features \cite{garg2022leveraging,baek2022agreement,duan2026preinterventionpredictionsparseautoencoder}, while recent monitor work studies retained safety representations and privacy guardrails \cite{Zhu2024Locking,das2026neurofilter}. Existing work therefore shows that activations contain safety-relevant signals, but does not systematically test whether frozen probes remain reliable after routine model updates.

\section{Experimental Setup}

\begin{table}[t]
\centering
\small
\setlength{\tabcolsep}{4pt}
\begin{tabular}{@{}p{0.36\linewidth}p{0.58\linewidth}@{}}
\toprule
\textbf{Component} & \textbf{Setting} \\
\midrule
Models & Gemma-2-2B-it, Qwen2.5-7B-Instruct \\
Monitors & harmfulness, high-stakes, privacy/PII, refusal-compliance \\
Layers & 4, 8, 12, 16, 20, 24 \\
Seeds & 0--4 \\
Probe & frozen linear probe \\
Completed update cells & 2520 \\
Main metrics & $\Delta$AUC, big-drop rate, operational failure \\
\bottomrule
\end{tabular}
\caption{Summary of models, monitors, layers, seeds, probe type, update families, and completed update cells used in the full-scale staleness evaluation.}
\label{tab:staleness-grid}
\end{table}

We evaluate two open-weight instruction-tuned models spanning a 3.5$\times$ parameter range: Gemma-2-2B-it \cite{team2024gemma} and Qwen2.5-7B-Instruct \cite{yang2025qwen3}. Both are loaded in bfloat16 as the base condition; all update conditions are derived from these checkpoints. The grid is model-agnostic by construction and supports drop-in substitution of further models.

\textbf{Monitors}. We study four binary safety-relevant monitoring tasks. Harmfulness uses WildGuardMix prompt-level harm labels \cite{han2024wildguard}; privacy/PII uses a balanced PII-benchmark-style dataset \cite{jha2026piibench} with personal-information positives and no-labeled-PII negatives; refusal-compliance uses XSTest \cite{rottger2023xstest}, contrasting safe prompts that resemble unsafe requests with genuinely unsafe prompts while preserving its natural size to avoid cross-source style confounds; and high-stakes is a controlled synthetic monitor with matched low- and high-stakes templates over the same topic domains. Each monitor uses up to 1,000 examples, balanced 500/500 by class when possible, with a 70/30 train/test split, exact-duplicate removal, and grouped splits for shared templates or pair identities.

\begin{table}[t]
\centering
\scriptsize
\setlength{\tabcolsep}{2pt}
\begin{tabular}{@{}p{0.34\linewidth}p{0.34\linewidth}p{0.22\linewidth}@{}}
\toprule
\textbf{Update family} & \textbf{Update types} & \textbf{Models covered} \\
\midrule
Quantization & bnb INT8, bnb NF4 & Gemma + Qwen \\
Weight-only quant & GPTQ, AWQ & Qwen only \\
Activation-aware quant & W8A8/SmoothQuant & Qwen only \\
LoRA & alpaca/general, safety, privacy & Gemma + Qwen \\
Merged LoRA & merged versions & Gemma + Qwen \\
QLoRA & safety QLoRA & Gemma + Qwen \\
\bottomrule
\end{tabular}
\caption{Update conditions and backend coverage. The designed grid contains twelve post-base update conditions.}
\label{tab:update-coverage}
\end{table}

\textbf{Update conditions}. We evaluate twelve post-base update conditions, with per-model backend coverage summarized in Table~\ref{tab:update-coverage}. Quantization includes bitsandbytes INT8 and NF4 \cite{dettmers2022llmint8,dettmers2023qlora}, weight-only GPTQ and AWQ \cite{frantar2023gptqaccurateposttrainingquantization,lin2026awqactivationawareweightquantization}, and activation-aware SmoothQuant-style W8A8 \cite{xiao2023smoothquant}. Fine-tuning includes three rank-16 LoRA adapters \cite{hu2022lora} for general instruction following, refusal/safety behavior, and privacy/PII-removal rewriting, their merged counterparts, and a QLoRA safety adapter served on an NF4-quantized base \cite{dettmers2023qlora}. Safety and privacy fine-tuning data are disjoint from the corresponding monitor datasets, with verbatim-overlap checks to avoid train-on-test leakage.

\textbf{Probes.} Monitors are implemented as linear logistic probes on residual-stream activations at the final non-padding token of the chat-formatted prompt, read at layers {4, 8, 12, 16, 20, 24} (all valid for both models). For each (model, monitor, layer) we train five probes on independent 85\% subsamples of the training split (seeds 0--4). Probes are trained only on base-model activations and are frozen thereafter: every update condition is evaluated by re-extracting activations from the updated model and scoring them with the unchanged probe. 

\begin{table}[t]
\centering
\scriptsize
\setlength{\tabcolsep}{2pt}
\begin{tabular}{@{}>{\raggedright\arraybackslash}p{0.30\linewidth}>{\raggedleft\arraybackslash}p{0.07\linewidth}>{\raggedleft\arraybackslash}p{0.18\linewidth}>{\raggedleft\arraybackslash}p{0.17\linewidth}>{\raggedleft\arraybackslash}p{0.18\linewidth}@{}}
\toprule
\textbf{Update family} & \textbf{$n$} & \textbf{Median $\Delta$AUC} & \textbf{Big-drop rate} & \textbf{Operational failure} \\
\midrule
Quantization & 480 & -0.0021 & 1.46\% & 0.00\% \\
Weight-only quant & 240 & -0.0059 & 3.33\% & 0.00\% \\
Activation-aware quant & 120 & -0.0062 & 0.83\% & 0.00\% \\
LoRA & 720 & -0.0374 & 43.33\% & 13.75\% \\
Merged LoRA & 720 & -0.0381 & 43.19\% & 13.75\% \\
QLoRA & 240 & -0.0571 & 53.75\% & 13.75\% \\
\bottomrule
\end{tabular}
\caption{Probe degradation summarized by update family using median $\Delta$AUC, big-drop rate, and operational-failure rate.}
\label{tab:update-family-degradation}
\end{table}

\begin{figure}[t]
\centering
\includegraphics[width=0.86\linewidth]{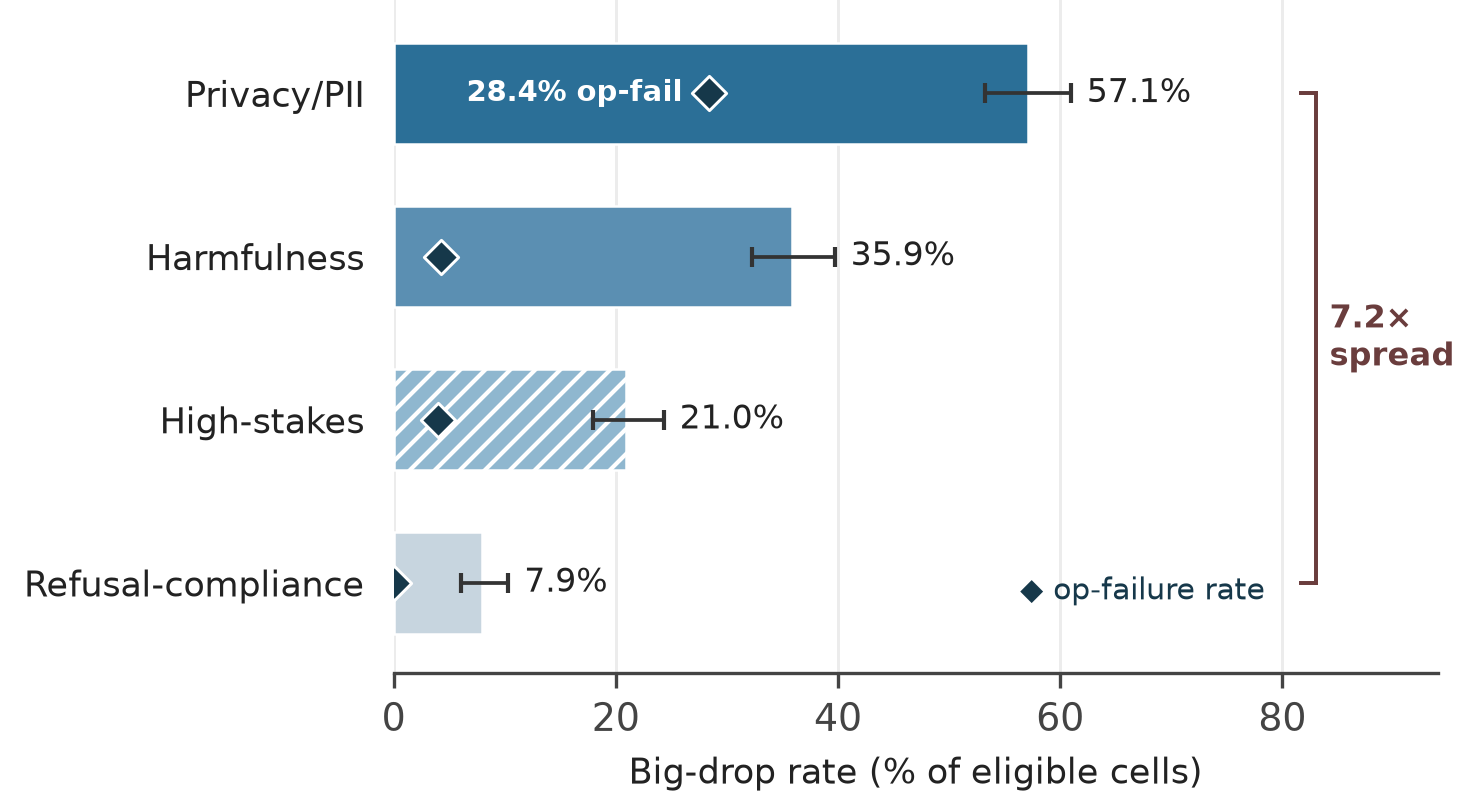}
\caption{\textbf{Monitor fragility is highly task-dependent.} Privacy/PII probes stale most often, while refusal-compliance probes are comparatively stable, producing a 7.2$\times$ spread under the same update distribution. Error bars show Wilson 95\% CIs; diamonds mark operational-failure rates.}
\label{fig:monitor-fragility}
\vspace{0.6em}
\captionsetup{type=table}
\scriptsize
\setlength{\tabcolsep}{2pt}
\resizebox{\linewidth}{!}{%
\begin{tabular}{@{}lrrrrrrr@{}}
\toprule
\textbf{Monitor} & \textbf{$n$} & \textbf{Median base AUC} & \textbf{Median $\Delta$AUC} & \textbf{Median $\Delta$ recall@1\% FPR} & \textbf{Median $\Delta$ recall@5\% FPR} & \textbf{Big-drop rate} & \textbf{Op. failure rate} \\
\midrule
Harmfulness & 630 & 0.9457 & -0.0240 & -0.1000 & -0.0867 & 35.87\% & 4.29\% \\
High-stakes & 630 & 0.9827 & -0.0142 & -0.1167 & -0.0694 & 20.95\% & 3.97\% \\
Privacy/PII & 630 & 0.8477 & -0.0886 & -0.0667 & -0.1377 & 57.14\% & 28.41\% \\
Refusal-compliance & 630 & 0.9928 & -0.0052 & -0.0333 & -0.0167 & 7.94\% & 0.00\% \\
\bottomrule
\end{tabular}%
}
\caption{Base probe quality and low-FPR degradation by monitor.}
\label{tab:monitor-fragility-metrics}
\end{figure}

\textbf{Metrics and grid.} Our primary quantity is the change in held-out AUC of the frozen probe between the base and updated model; we additionally record recall at 1\% and 5\% false-positive rate and expected calibration error. The full grid is 2 models $\times$ 4 monitors $\times$ 12 update conditions $\times$ 6 layers $\times$ 5 seeds = 2,880 designed update cells, of which 2,520 completed (the remainder are the logged Gemma cells lacking public quantized checkpoints); all completed cells passed the base-quality eligibility gate fixed before the full-scale run (Table \ref{tab:staleness-grid})

\section{Benchmarking Activation Monitor Staleness}

In this section, we systematically benchmarks activation-monitor staleness under routine model updates. We first define the frozen-probe evaluation protocol and degradation criteria, then show that failures are structured by update family and monitor type rather than uniformly distributed.

\paragraph{Activation-Monitor Persistence.}
Let $M$ be a base language model and let $\phi_\ell^M(x) \in \mathbb{R}^d$ denote the residual-stream activation at the final non-padding token of input $x$ after decoder block $\ell$. A monitor is a dataset $D = \{(x_i, y_i)\}$ of prompts with binary safety-relevant labels together with a probe $h$ trained on $\{(\phi_\ell^M(x_i), y_i) : i \in D_{\mathrm{train}}\}$. An update is an operator $U$ mapping $M$ to an updated model $M' = U(M)$ with the same architecture and tokenizer interface, so layer indices remain aligned, but altered weights or numerics---quantization, adapter insertion, adapter merging, or quantized-base adaptation. The persistence question is whether $h$, trained on $\phi_\ell^M$ and frozen, remains accurate when composed with $\phi_\ell^{M'}$.

\paragraph{Benchmark Cells.}
The unit of evaluation is a cell $c = (\text{model}, \text{monitor}, \text{update}, \text{layer}, \text{seed})$, where the seed indexes the probe-training subsample. For each cell we compute the base score $A_{\mathrm{base}}(c) = \mathrm{AUC}(h \circ \phi_\ell^M; D_{\mathrm{test}})$ and the post-update score $A_{\mathrm{post}}(c) = \mathrm{AUC}(h \circ \phi_\ell^{M'}; D_{\mathrm{test}})$ on the identical held-out split, and define
\[
\Delta\mathrm{AUC}(c) = A_{\mathrm{post}}(c) - A_{\mathrm{base}}(c).
\]
Before the full-scale run, we fixed three derived criteria: $\mathrm{big\_drop}(c) = \mathbf{1}[\Delta\mathrm{AUC}(c) \leq -0.05]$, marking degradation large enough to change operational conclusions; $\mathrm{operational\_failure}(c) = \mathbf{1}[A_{\mathrm{post}}(c) < 0.70]$, marking a probe that is no longer usable regardless of its starting point; and $\mathrm{eligible}(c) = \mathbf{1}[A_{\mathrm{base}}(c) \geq 0.75]$, restricting analysis to cells where a meaningful monitor existed to degrade. All aggregate statistics (median $\Delta\mathrm{AUC}$, big-drop rate, operational-failure rate) are computed over eligible cells.

\subsection{Update Family Determines Monitor Staleness}

Frozen monitors frequently go stale after fine-tuning, but rarely after quantization. Table~\ref{tab:update-family-degradation} gives the family-level picture. The three quantization families are nearly flat: median \(\Delta\)AUC ranges from \(-0.0021\) (bitsandbytes INT8/NF4, \(n = 480\)) through \(-0.0059\) (weight-only GPTQ/AWQ, \(n = 240\)) to \(-0.0062\) (activation-aware W8A8, \(n = 120\)), big-drop rates range from \(0.83\%\) to \(3.33\%\), and none of the \(840\) quantization-family cells produces an operational failure.

The fine-tuning families are an order of magnitude worse on every statistic: median \(\Delta\)AUC of \(-0.0374\) (LoRA, \(n = 720\)), \(-0.0381\) (merged LoRA, \(n = 720\)), and \(-0.0571\) (QLoRA, \(n = 240\)); big-drop rates of \(43.33\%\), \(43.19\%\), and \(53.75\%\); and a \(13.75\%\) operational-failure rate in each family. Confidence intervals preserve the same separation: aggregate quantization has a big-drop rate of \(1.90\%\) with Wilson \(95\%\) CI \([1.18, 3.07]\) and no operational failures, while LoRA, merged LoRA, and QLoRA have big-drop rates of \(43.33\%\) \([39.76, 46.98]\), \(43.19\%\) \([39.62, 46.84]\), and \(53.75\%\) \([47.43, 59.95]\), respectively.

The family medians sit near the big-drop threshold while nearly half the cells cross it, indicating a heavy left tail concentrated in particular monitor \(\times\) layer pockets rather than a uniform shift; consistent with this, operational failures recur at an identical rate across all three fine-tuning families, suggesting the same vulnerable cells fail under every adapter. We confirm this overlap directly: LoRA and merged LoRA have near-identical failing-cell sets, with Jaccard overlap \(0.980\) for big drops and \(0.982\) for operational failures. QLoRA overlaps less but still substantially with the LoRA families (\(0.735\)--\(0.750\) for big drops), suggesting both shared vulnerable monitor/layer pockets and additional quantized-adaptation-specific failures.

\begin{figure}[t]
\centering
\includegraphics[width=\linewidth]{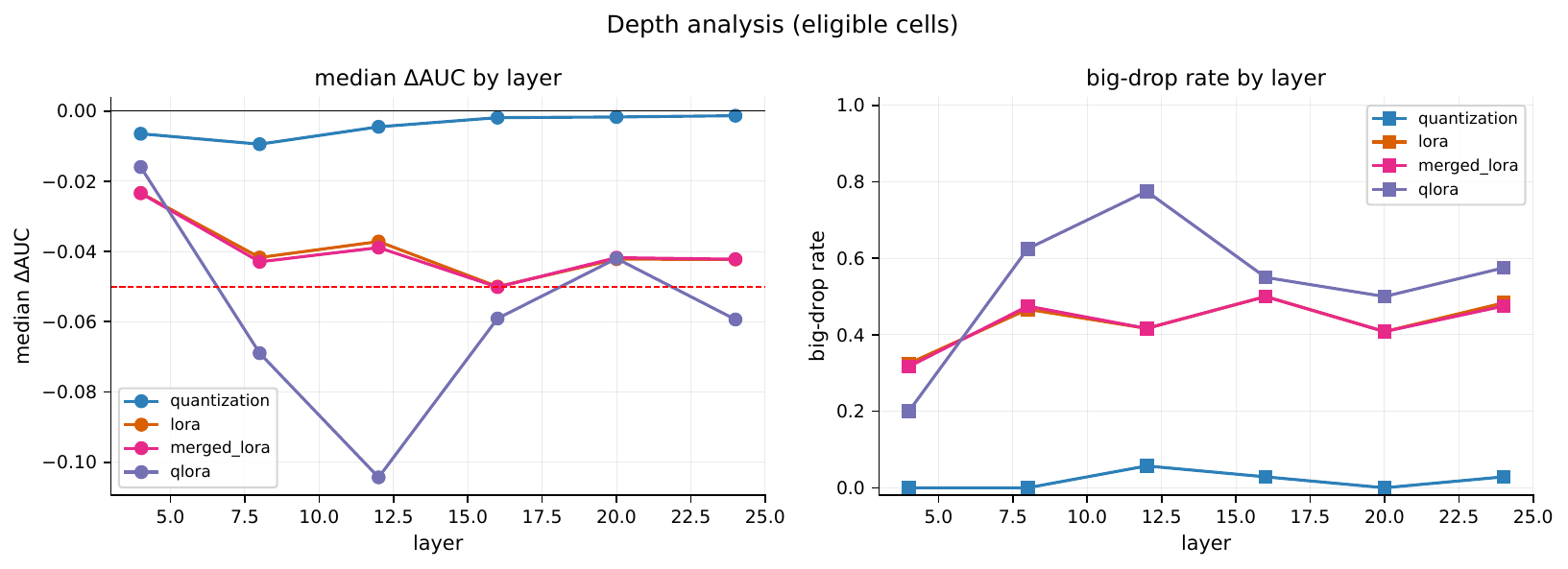}
\caption{\textbf{Fine-tuning degrades probes across depth; quantization does not.} Median $\Delta$AUC (left) and big-drop rate (right) by layer and update family. Quantization stays near-flat at all depths; LoRA, merged-LoRA, and QLoRA degrade across most layers, QLoRA worst at middle layers.}
\label{fig:layer-depth}
\end{figure}

Figure~\ref{fig:layer-depth} characterizes this layer axis: quantization remains near-flat at every measured depth, whereas LoRA-style updates degrade probes across most layers, with the strongest QLoRA degradation concentrated around middle layers.

Runtime and merged LoRA produce nearly identical degradation in paired comparison. Across all matched runtime/merged cells, the median paired \(\Delta\)AUC difference is \(0.0000\), the pooled Wilcoxon test does not show evidence of a reliable overall shift (\(p = 0.1369\)), and big-drop/operational-failure discordance is rare: only \(2\) versus \(1\) discordant big-drop cells and \(1\) versus \(1\) discordant operational-failure cells, with exact discordance tests \(p = 1.0\). This supports the merged conditions' intended role as a numerical control: staleness is attributable to the weight update itself, not the serving path.

QLoRA, which combines an NF4 base with adaptation, is the most damaging condition in the grid. We note that GPTQ, AWQ, and W8A8 were available only for Qwen2.5-7B, so comparisons among quantization families are within-model; the quantization-versus-fine-tuning contrast holds within both models. The qualitative split is also insensitive to the exact big-drop cutoff: sweeping the threshold from \(-0.03\) to \(-0.07\), aggregate quantization remains below \(7\%\) big-drop rate, while every LoRA-family update remains above \(35\%\).

\subsection{Monitor Tasks Differ Sharply in Fragility}

Fragility varies strongly across monitor type. Figure~\ref{fig:monitor-fragility} shows big-drop rates by monitor over $n = 630$ eligible cells each: privacy/PII 57.14\%, harmfulness 35.87\%, high-stakes 20.95\%, and refusal-compliance 7.94\%---a $7\times$ spread between the most and least fragile monitor under the identical update distribution. Privacy/PII is the dominant failure case, accounting for 360 of the 768 big drops. Table \ref{tab:monitor-fragility-metrics} shows that this fragility also appears at low false-positive operating points: privacy/PII has the largest median recall loss at \(5\%\) FPR and the highest operational-failure rate, while refusal-compliance remains comparatively stable. Refusal-compliance is the most stable despite being trained on the smallest dataset, though its smaller test split makes its rate estimate the noisiest of the four. We reiterate that high-stakes is a controlled synthetic monitor; its absolute rate should be read as a diagnostic on a clean construct rather than a field estimate. The update-family conclusion does not rely on this synthetic construct: after excluding high-stakes, the aggregate quantization big-drop rate remains low at \(2.54\%\) with no operational failures, while LoRA, merged LoRA, and QLoRA rise to \(47.59\%\), \(47.59\%\), and \(58.89\%\) drop rates.

\begin{table}[t]
\centering
\scriptsize
\setlength{\tabcolsep}{2pt}
\resizebox{\linewidth}{!}{%
\begin{tabular}{@{}>{\raggedright\arraybackslash}p{0.27\linewidth}>{\raggedleft\arraybackslash}p{0.15\linewidth}>{\raggedright\arraybackslash}p{0.52\linewidth}@{}}
\toprule
\textbf{Check} & \textbf{$n$} & \textbf{Main result} \\
\midrule
Retrained-probe recovery & 42 cells & Median recovery fraction = \textbf{0.981} \\
Adapter-seed variance & 3 seeds & Max median-$\Delta$AUC range = \textbf{0.024} \\
Response-side monitor & 12 cells & Response-side $\Delta$AUC = \textbf{-0.0817}; prompt-side = \textbf{-0.0302} \\
\bottomrule
\end{tabular}%
}
\setcounter{table}{4}
\caption{Mechanism and robustness analyses across recovery, adapter-seed variance, and response-side monitoring.}
\label{tab:mechanism-robustness-analyses}
\end{table}
\setcounter{table}{3}

\subsection{Mechanism and Robustness Analyses}

We next test whether staleness reflects signal erasure, adapter-seed noise, or our use of prompt-final-token activations. Retrained probes on updated activations recover most lost performance (Table~\ref{tab:mechanism-robustness-analyses}), showing that the monitored signal usually survives but shifts away from the frozen base-probe direction. Independent safety-LoRA adapters produce similar degradation, ruling out a single anomalous adapter run, and response-side probes show comparable or larger drops, ruling out a prompt-token artifact. These checks support a representational-drift interpretation: fine-tuning preserves the monitored distinction but moves the internal readout direction enough to stale a frozen probe.

The monitor ordering also reveals an asymmetry: safety LoRA leaves refusal-compliance comparatively stable, while privacy-targeted tuning most strongly degrades privacy/PII probes. QLoRA is additionally more damaging than NF4 quantization alone, suggesting that the risk comes from adaptation on a quantized base rather than quantization by itself.

\section{Predicting Which Monitors Need Revalidation}

This section tests whether activation-monitor staleness is predictable before post-update evaluation. We use only pre-deployment features available before revalidating the updated model, then compare learned predictors against strong lookup-table baselines to determine whether degradation can be triaged rather than discovered exhaustively.

\paragraph{Staleness Prediction.}

Beyond characterizing degradation, we ask whether it is predictable before post-update evaluation. The predictor maps $z(c)$ to $\Delta\mathrm{AUC}(c)$ for regression or $\mathrm{big_drop}(c)$ for classification. Every feature in $z(c)$ is observable pre-deployment: cell metadata, base-probe quality, derived fragility quantities, and base-model probe/activation geometry. These include model, monitor, update, update family, layer, normalized depth, base AUC, ECE, recall at fixed FPR, eligibility margin, recall gaps, depth indicators, update-intent $\times$ monitor terms, family $\times$ early-layer terms, probe weight norm, class-centroid distance, between/within variance ratio, training-margin statistics, and score entropy. These form four nested feature sets---metadata-only $\subset$ base-metrics $\subset$ fragility $\subset$ fragility+geometry---used to attribute predictive power.

\paragraph{Baselines.}
Because grid structure alone can explain substantial variation, we compare against eight group-mean lookup baselines: global mean; means by update, update family, update $\times$ layer, update $\times$ monitor, update $\times$ model, update $\times$ layer $\times$ monitor, and update $\times$ layer $\times$ monitor $\times$ model. If a test key is unseen in training, as in leave-update-out, the lookup backs off to the largest seen subset of its key columns before falling back to the global mean. The claim of interest is the margin of the best learned predictor over the best of these baselines.

\paragraph{Evaluation Schemes.}
We evaluate predictors under stratified 5-fold cross-validation as an interpolation reference and four extrapolation schemes: leave-update-out (12 folds), leave-monitor-out (4), leave-model-out (2), and leave-update-family-out (6). The primary metric is mean within-fold Spearman correlation between predicted and observed $\Delta\mathrm{AUC}$; pooled correlations are secondary because heterogeneous folds can inflate rank correlations. Predictors include elastic net, random forest, extra trees, and histogram gradient boosting. For each scheme, we compare the best predictor $\times$ feature-set configuration against the best of the eight baselines, using a symmetric best-vs-best comparison over fixed model families. Success requires beating the best baseline by at least $+0.05$ mean within-fold Spearman in at least two of the four leave-out schemes.

\subsection{Degradation Is Predictable Beyond Baselines}

\begin{table}[t]
\centering
\scriptsize
\setlength{\tabcolsep}{2pt}
\resizebox{\linewidth}{!}{%
\begin{tabular}{@{}lrrrrr@{}}
\toprule
\textbf{Feature set} & \textbf{Spearman} & \textbf{Pearson} & \textbf{MAE} & \textbf{RMSE} & \textbf{$\Delta\rho$} \\
\midrule
Metadata & 0.669 & 0.664 & 0.026 & 0.042 & --- \\
+ Base metrics & 0.678 & 0.681 & 0.025 & 0.040 & +0.009 \\
+ Fragility & 0.681 & 0.678 & 0.025 & 0.040 & +0.003 \\
+ Geometry/artifacts & 0.703 & 0.713 & 0.024 & 0.038 & +0.022 \\
\bottomrule
\end{tabular}%
}
\caption{Predictor feature-set ablation. Adding base metrics and fragility features gives modest gains, while geometry/artifact features provide the largest incremental improvement.}
\label{tab:predictor-feature-ablation}
\end{table}

\begin{figure}[t]
\centering
\includegraphics[width=0.78\linewidth]{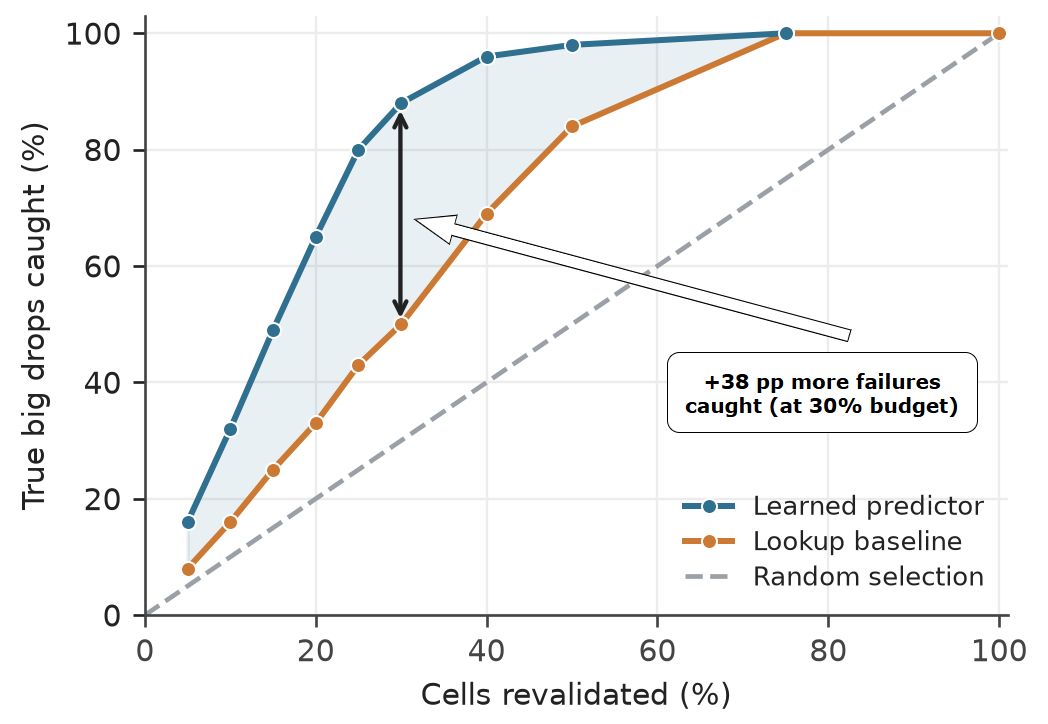}
\caption{\textbf{Prediction triages the revalidation budget efficiently.} Fraction of true big-drops caught vs. fraction of cells revalidated, ranking cells by predicted big-drop risk. The learned predictor catches failures far faster than the lookup baseline or random selection — up to +38 points more at 30\% budget.}
\label{fig:predictor-budget-curve}
\vspace{0.6em}
\captionsetup{type=table}
\scriptsize
\setlength{\tabcolsep}{2pt}
\begin{tabular}{@{}>{\raggedright\arraybackslash}p{0.33\linewidth}>{\raggedleft\arraybackslash}p{0.23\linewidth}>{\raggedleft\arraybackslash}p{0.20\linewidth}>{\raggedleft\arraybackslash}p{0.12\linewidth}@{}}
\toprule
\textbf{CV scheme} & \textbf{Best baseline Spearman} & \textbf{Best ML Spearman} & \textbf{$\Delta$} \\
\midrule
Leave-update-out &
0.6324 &
\textbf{0.7239} &
\textbf{\textcolor{ForestGreen}{+0.0915}} \\
Leave-monitor-out &
0.6165 &
\textbf{0.6694} &
\textbf{\textcolor{ForestGreen}{+0.0529}} \\
Leave-model-out &
0.6433 &
0.6690 &
\textcolor{gray}{+0.0257} \\
Leave-family-out &
0.5813 &
\textbf{0.7174} &
\textbf{\textcolor{ForestGreen}{+0.1361}} \\
\bottomrule
\end{tabular}
\setcounter{table}{6}
\caption{Prediction performance against lookup-table baselines. Green bold margins exceed the prespecified $+0.05$ success threshold.}
\label{tab:prediction-cv-schemes}
\end{figure}

Monitor degradation is predictable beyond lookup-table baselines. Table~\ref{tab:prediction-cv-schemes} reports the regression results under the four extrapolation schemes. The group-mean baselines are strong, reaching mean within-fold Spearman of 0.58--0.64, which underscores that raw predictor correlations would be uninformative; the claim rests on the margin over the best baseline. The learned predictor exceeds the fixed $+0.05$ margin in three of four schemes: leave-update-out ($+0.0915$, $0.6324 \to 0.7239$), leave-monitor-out ($+0.0529$, $0.6165 \to 0.6694$, narrowly clearing the margin), and leave-update-family-out ($+0.1361$, $0.5813 \to 0.7174$)---the last being the hardest extrapolation, in which an entire update family unseen in training must be ranked. Leave-model-out does not clear the margin ($+0.0257$, $0.6433 \to 0.6690$); with only two models, model-specific structure cannot transfer and the update $\times$ layer $\times$ monitor lookups already capture most cross-model regularity. The locked criterion ($\geq +0.05$ in $\geq 2$ of 4 schemes) is met. Table~\ref{tab:predictor-feature-ablation} ablates the predictor inputs: metadata is already informative, but artifact/geometry features provide the largest incremental gain, indicating that representation-level properties help predict which monitors become stale. Practically, this ranking translates into efficient triage: ordering monitor $\times$ layer cells by predicted big-drop risk catches far more true failures per unit of revalidation budget than the lookup-table or random-selection baselines.

\begin{figure*}[t]
\centering
\begin{minipage}[t]{0.48\textwidth}
\vspace{0pt}
\centering
\includegraphics[width=\linewidth,height=0.135\textheight]{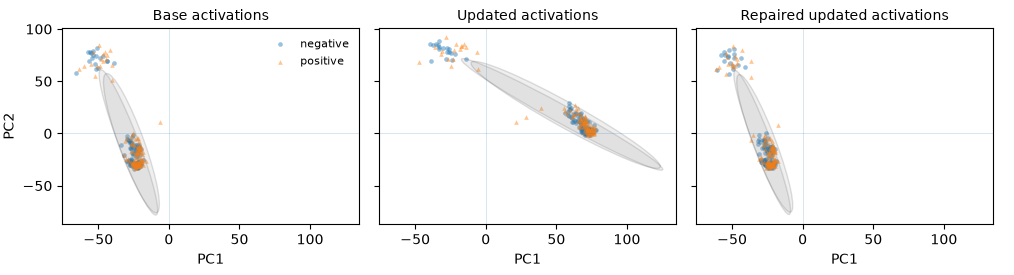}
\caption{ \textbf{Staleness reflects representational drift.} PCA projections show that updated activations retain class structure but shift away from the base activation geometry. Label-free realignment maps the updated activations back toward the base space, restoring compatibility with the frozen probe.}
\label{fig:representational-drift-pca}
\end{minipage}
\hfill
\begin{minipage}[t]{0.48\textwidth}
\vspace{0pt}
\centering
\includegraphics[width=0.78\linewidth]{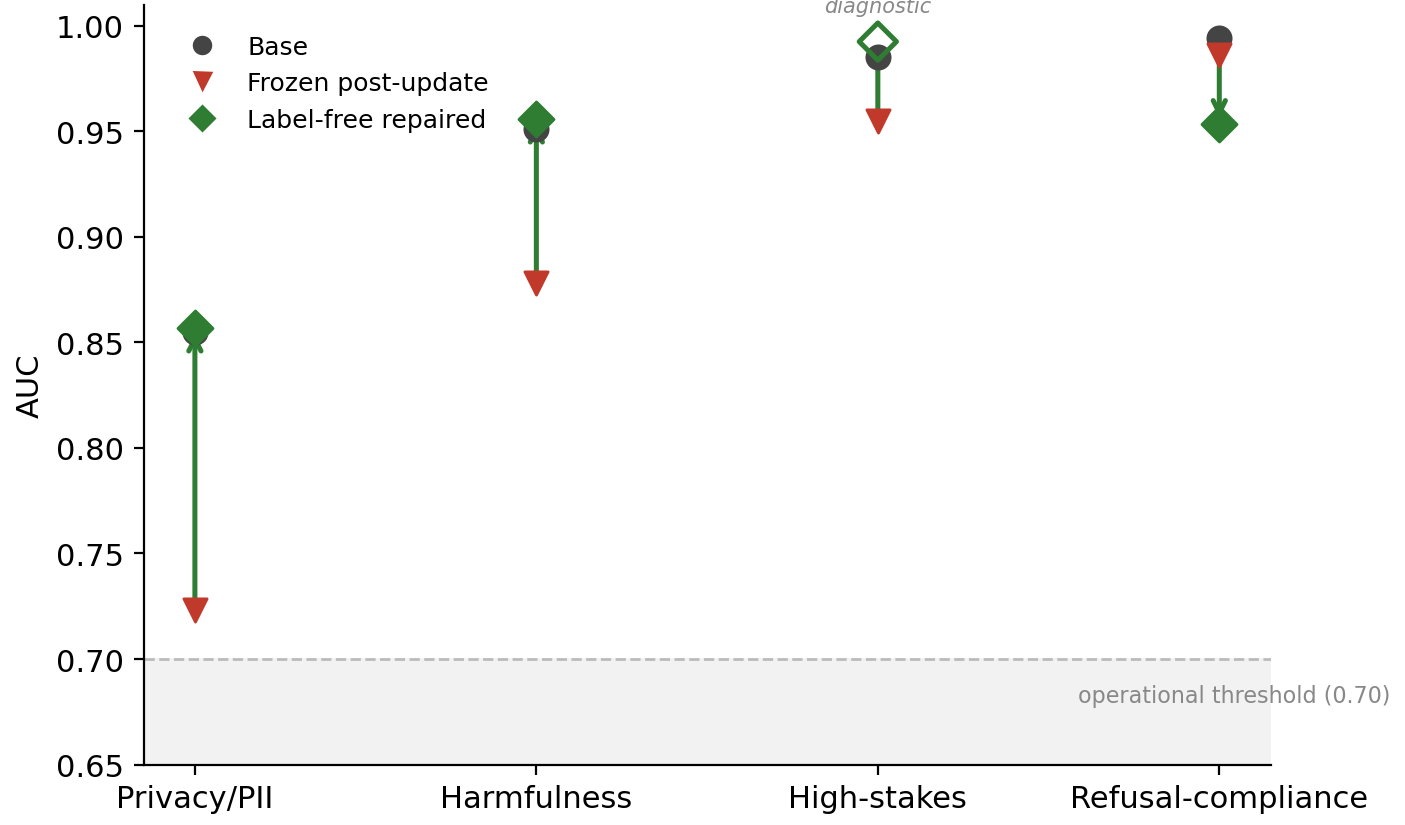}
\caption{\textbf{Learned triage catches failures earlier than lookup baselines.} Ranking cells by predicted big-drop risk recovers true failures much faster than the strongest lookup baseline or random selection, allowing revalidation budget to focus on high-risk monitor--update cells.}
\label{fig:repair-recovery-by-monitor}
\end{minipage}
\end{figure*}

\section{Repairing Stale Monitors}

This section tests whether activation-monitor staleness is repairable rather than permanent. We evaluate a cost-ordered repair ladder and show that paired unlabeled base/update activations can often realign updated representations enough for the original frozen probe to recover.

\paragraph{Label-Free Realignment and Repair Taxonomy.}

To repair stale monitors without new labels, we fit one alignment map per model, update, and layer from paired unlabeled base/update activations, then evaluate the frozen probe on mapped updated activations, $h(A\phi^{M'}_{\ell}(x))$. We compare a cost-ordered ladder—identity, label-free realignment, score calibration, few-label heads, and full retraining—and assign each cell the cheapest rung that recovers at least half the lost AUC or restores AUC above 0.70. We evaluate this on a complete 216-cell targeted grid with disjoint alignment prompts and random-orthogonal and shuffled-pair controls.

\subsection{Stale Monitors Can Be Repaired Without Labels}
On the complete 216-cell grid, 105 cells were not stale enough to require repair and the remaining 111 repair-relevant cells were resolved by the cheapest rung, label-free realignment; no cell required score calibration, a few-label head, or full retraining. Recovery tracked fragility. Privacy/PII was the most update-sensitive monitor (51 of 54 cells repaired; frozen $\Delta$AUC $-0.13$ to $-0.15$ across the three updates) yet was recovered nearly completely, with best label-free AUC $0.853$--$0.861$ against base $0.855$ (recovery fraction $\approx 0.92$--$1.02$). Harmfulness degraded mainly under safety-style updates (27 of 54 repaired; stable under \texttt{lora\_privacy} with $\Delta$AUC $-0.017$) and was likewise restored to $0.954$--$0.958$ against base $0.951$ ($\approx 0.99$--$1.11$). Refusal-compliance was the most stable (9 of 54 repaired; median $\Delta$AUC $-0.001$ to $-0.010$) and high-stakes showed moderate drift (24 of 54 repaired; $\Delta$AUC $-0.013$ to $-0.045$) that we read as diagnostic given its synthetic construction. Random-orthogonal and shuffled-pair controls produced large negative recovery throughout, ruling out trivial fits. We note that the taxonomy does not discriminate among rungs above label-free: because the cheapest rung sufficed in every repair-relevant cell, the calibration, few-label, and retraining branches are validated as code paths but never become cheapest. Together, the monitor-wise pattern supports recoverable representational drift rather than signal erasure.

\section{Discussion}

\textbf{Deployment implications}. In our evaluated settings, the operational asymmetry is stark: fine-tuning events --- including parameter-efficient ones, regardless of whether adapters are served at runtime or merged --- degraded 43--54\% of monitor cells past the big-drop threshold and rendered roughly one in seven unusable, while 840 quantization cells produced no operational failures. This suggests that any weight-adaptation event in a deployment pipeline should trigger monitor revalidation as a matter of course, with label-free realignment as a cheap first repair and labeled probe retraining reserved for monitors that remain stale after repair; quantization appears lower-risk, but its nonzero big-drop rate (up to 3.33\% for weight-only schemes) argues for cheap spot checks rather than exemption. The predictor's role is triage, not certification: because all of its features are available before any post-update evaluation, it can rank monitor $\times$ layer cells by expected degradation and direct a limited revalidation budget toward the likely failures first --- an early-warning system that complements, and must not replace, actual re-measurement.

\textbf{Why quantization is flat but fine-tuning is not.} The two update families differ in what they are constrained to preserve. Post-training quantization is designed to preserve the input--output function, and our results suggest that this often also leaves the probe-relevant activation geometry nearly unchanged. Fine-tuning has no such constraint: even low-rank updates can rotate internal coordinates enough that a frozen base probe misreads the updated representation, while the monitored signal remains linearly recoverable. This also explains why QLoRA is riskier than NF4 alone: the quantized base is relatively benign until adaptation moves the representation. Thus, we interpret the split as consistent with function-preserving compression versus representation-moving adaptation, while noting that the benchmark measures where probes fail rather than fully explaining why.

\textbf{Why privacy/PII monitors may be especially fragile.} The privacy construct is deliberately narrow --- the presence of concrete PII values in otherwise matched text --- and such fine-grained, surface-proximal features may occupy lower-variance directions that fine-tuning displaces more easily than broad semantic axes like harmfulness. This susceptibility is plausibly compounded by intent: two of our seven fine-tuning conditions directly retrain the model's handling of PII-bearing text, so part of the degradation likely reflects representational change in exactly the subspace the monitor reads.

\section{Conclusion}

We asked whether the implicit contract of activation monitoring holds --- that a probe trained on base-model activations keeps reading the same safety-relevant geometry after routine model-side updates --- and the answer is sharply structured rather than uniform. Function-preserving quantization leaves frozen monitors largely intact, while parameter-efficient fine-tuning frequently stales them, and the effect varies sharply by monitor. The key point is that this staleness is often drift rather than erasure: the monitored signal usually survives the update, but the frozen readout direction silently falls out of alignment with it. This failure is invisible without re-measurement, predictable in advance from pre-deployment features, and sometimes cheaply repairable by label-free activation realignment. Practically, weight adaptation should trigger monitor revalidation by default, with prediction triaging which monitors to check first, label-free repair attempted when possible, and labeled retraining reserved for monitors that remain stale after repair.

\section*{Limitations}

We evaluate two open-weight instruction-tuned models so cross-model generality is not established. Our monitors are incomplete proxies for deployment safety. High-stakes is a synthetic diagnostic. GPTQ, AWQ, and SmoothQuant-style W8A8 checkpoints were available only for Qwen2.5-7B. Finally, the study is behavioral, not mechanistic: we do not explain which internal directions move. Future work should study mechanisms, and broader models.


\end{document}